# Conceptual Modeling of Time for Computational Ontologies

**Sabah Al-Fedaghi**
*sabah.alfedaghi@ku.edu.kw*
Computer Engineering Department, Kuwait University, Kuwait

**Summary**
To provide a foundation for conceptual modeling, ontologies have been introduced to specify the entities, the existences of which are acknowledged in the model. Ontologies are essential components as mechanisms to model a portion of reality in software engineering. In this context, a model refers to a description of objects and processes that populate a system. Developing such a description constrains and directs the design, development, and use of the corresponding system, thus avoiding such difficulties as conflicts and lack of a common understanding. In this cross-area research between modeling and ontology, there has been a growing interest in the development and use of domain ontologies (e.g., Resource Description Framework, Ontology Web Language). This paper contributes to the establishment of a broad ontological foundation for conceptual modeling in a specific domain through proposing a workable ontology (abbreviated as TM). A TM is a one-category ontology called a thimac (*thi*ngs/*ma*chines) that is used to elaborate the design and analysis of ontological presumptions. The focus of the study is on such notions as change, event, and time. Several current ontological difficulties are reviewed and remodeled in the TM. TM modeling is also contrasted with time representation in SysML. The results demonstrate that a TM is a useful tool for addressing these ontological problems.

***Keywords:***
*Ontology; computational ontology; conceptual model; SysML; engineering requirements; modeling time*

## 1. Introduction

According to Merrill [1], “ontologies, serve as essential components in the engine of contemporary science.” Any ontological matter refers to a concern of “What is there?” and “what is it that there is not?” [2]. A sample of ontological positions involves refusing to recognize certain entities and entities of a certain kind [2]. According to [3], ontology is “the science of what is, of the kinds and structures of objects, properties events, processes, and relations in every area of reality.” Although ontology is a philosophical discipline, applied ontology seems to break off as a special science [1]. Applied ontology “is concerned with building a ‘conceptual model’ of what it means for something to exist” [1].

### 1.1 Ontologies and Conceptual Modeling

To provide a foundation for conceptual modeling, ontologies have been introduced to specify the entities, the existences of which are acknowledged in the model [4]. Conceptual modeling involves “representing aspects of the world for the purpose of understanding and communication [, and hence] the contribution of a conceptual modeling notation rests in its ability to promote understanding about the depicted reality among human users” [5]. Conceptualization refers to abstracting a given portion of reality that “exists beyond our concepts” [6], i.e., entities and processes that “exist” in the *domain* of the modeled system (note that *what there* may not be *what exists* – see [7]). It is also stated that ontology is “an explicit specification of a conceptualization” [8].

In the field of conceptual modeling, there has been a growing interest in the development and use of domain ontologies. Domain ontology represents concepts that belong to a portion of the world (e.g., biology). Domain ontologies have become essential for research in areas such as machine learning, the Internet of Things, robotics, and natural language. Ontologies are intended to play a significant role in “facilitating seamless information processing and interoperability among applications” [8].

Ontology is also a matter of inquiry, development, and application in software engineering, because of the need to categorize and structure entities and concepts of interest in information systems [9]. Ontological research appears in various fields of computer science [1] (e.g., (Resource Description Framework, Ontology Web Language). Computational ontologies are becoming one of the most pervasive forms of emerging scientific media [10] and stand to revolutionize entire industries and domains of social life [11][12]. According to Kishore et al. [9], computational and philosophical ontology differ in at least two ways:

1. Computational ontology is an academic pursuit to gain knowledge on reality. Computational ontologies add the goal of being implemented and used in the pursuit of other pragmatic objectives in a specific application.
2. Philosophical ontology deals with all reality in the entire universe of discourse, whereas computational ontology deals with only the “reality of interest” and in only a bounded (limited) universe of discourse [9].

### 1.2 About this Paper

One of the main advantages of researching the notion of ontology in conceptual modeling lies in bringing clarity and directionality through highlighting the nature of whatever it is that one is attempting to model. Additionally, ontological analysis allows the identification of inconsistencies and other inadequacies in modeling through examining ontological assumptions of different methodologies. This contributes to clearing the ground so that substantive modeling can advance more productively than would otherwise be the case. In software engineering, *an* ontology models a system by describing things and processes that populate it. This constrains and directs the design, development, and use of the software system, thus avoiding such difficulties as conflicts and lack of a common understanding of requirements among users and modelers.

This paper contributes to the description of a broad ontological foundation in conceptual modeling for software engineering. The aim of applying the paper is twofold.

- In this paper, selected current ontological difficulties are reviewed and remodeled in the TM, and TM-based modeling is demonstrated as a useful tool for addressing these ontological problems. For example, incorporating time in SysML is examined and contrasted with incorporating time in TM modeling.
- The TM model provides a different way of modeling than the prevailing paradigm in software engineering (e.g., UML, ER, etc.). Accordingly, the TM approach must be investigated from all dimensions (e.g., ontologically and semantically), in addition to being applied to various applications. This paper is part of research aimed at developing an ontological foundation for the TM model, specifically regarding the meaning of the notions of change, event, and their relationship to time.

In general, the issues in this context "are not of a purely technical nature (that could be addressed within computer or information science), but rather involve fundamental questions concerning the relation of one conceptual scheme or ontology to another, how concepts should be characterized, and how two concepts may be related to one another if they appear in disparate complex systems [1].

The next section contains a brief look at ontologies in conceptual modeling because there are very recent reviews on the topic, such as in *ACM Computing Surveys* [8]. Section 3 presents an enhanced summary of the TM model. The example in section 3 is a new contribution. Section 4 examines static and dynamic modeling through studying changes, events, and time. Section 5 contrasts time-based modeling in SysML and TM.

## 2. Related Works

There has been active research for more than 20 years in domain ontology resulting in a very rich area of research. A recent article in *ACM Computing Survey*s [8] gives a comprehensive survey of topics related to ontologies: e.g., [13], [14], and [15]. According to the article [8], "Identifying a suitable ontology for a given task is nontrivial because ontologies are implemented using a variety of languages, methodologies, and platforms. Effective tools are thus needed to adequately address the ontology selection and evaluation problem."

Guizzardi et al. [16] includes a review of more related study focusing on developing ontological foundations for conceptual modeling. The paper discusses the development of the conceptual modeling language and a number of methodological and computational tools. The work describes developing ontological bases for conceptual modeling and organized around a unified foundational ontology. Their ontological theory is based on *a* four-category ontology [17]. Additionally, the problems presented in the previous section of this paper were discussed in an earlier line of this research, including [18] and [19].

## 3. Thinging Model Theory

This section briefly reviews TM-modeling notions with ontology-related enhancements. A more elaborate discussion of the TM model foundation can be found in [20-29].

### 3.1 Basics of the Thinging Machine Modeling

Ontology refers to the *categorical* structure of reality, which is typically hierarchical. All concepts belong to a category [30]. Since Aristotle, it has been assumed that things belong to fundamentally different ontological categories [30]. Different ontologies may be distinguished by the number of their categories. An example of "four-category" ontology consists of objects, kinds, attributes, and modes [17], whereas a three-category ontology has the basic categories of entities, processes, and states [31]. A two-category ontology posits only objects and universals, and a well-known one-category ontology includes only the so-called *tropes*. According to Paul [32],

> One category ontologies are deeply appealing, because their ontological simplicity gives them an unmatched elegance and sparseness… We don't need a fundamental categorical division between particulars, individuals, or spacetime regions and their properties, nor do we need a fundamental categorical division between things, individuals, or bearers and the qualities "borne" by them… Ontologies that postulate multiple fundamental categories assign excess structure to the beast of reality, making a mess of the carving.

### 3.2 Basic TM Model Constructs

The TM modeling is based on a one category called thimacs (*th*ings/*ma*chines), which is denoted by Δ. The Δ has a dual mode of being: the machine side, denoted as M, and the thing side, denoted by T (see Fig. 1). Thus, Δ = (M. T).

The notion of T (Thing) relies more on Heidegger's [33] notion of "things" than it does on *objects*, the latter being a very popular notion in computer science (e.g., object-oriented modeling). The term "machine" refers to a special abstract thinging machine – see Fig. 2, which shows a basic complete M. A machine can be a subdiagram of Fig. 2 (e.g., only create and process), or it can be a complex of these machines. M is built under the postulation that it performs five generic actions – creating, processing (altering), releasing, transferring, and receiving – or a subset or complex of these actions. A thing is created, processed, released, transferred, and/or received, whereas a machine creates, processes, releases, transfers, and/or receives things.

The five actions (also called stages) in Fig. 2 form the foundation for the Δ-based modeling. Among the five stages, flow (a solid arrow in Fig. 2) signifies conceptual movement from one machine to another or among the machine's stages. The stages can be described as follows.

- *Arrival*: A thing reaches a new machine.
- *Acceptance*: A thing is permitted to enter the machine. If arriving things are always accepted, then arrival and acceptance can be combined into a stage of "receiving." For simplicity, this paper's examples assume that a receive stage exists.
- *Processing* (alteration): A thing undergoes a transformation that modifies it without creating a new thing.
- *Release*: A thing is marked as ready to be transferred outside of the machine.
- *Transference*: A thing is input or output outside of/in the machine.

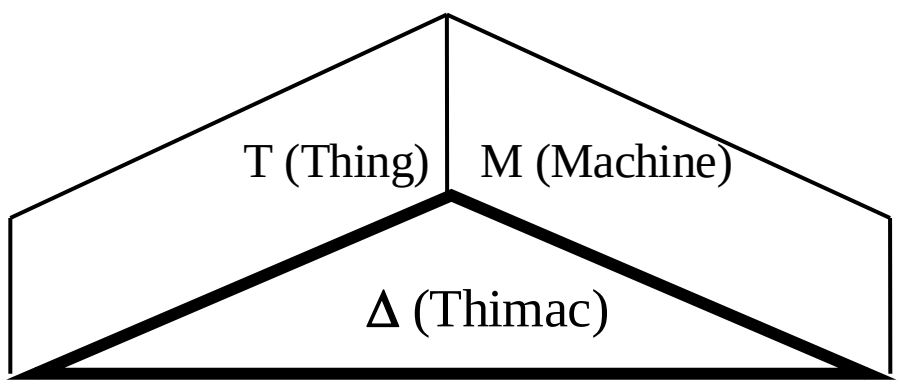

Fig. 1. The thimac has a dual mode of being a thing and a machine.

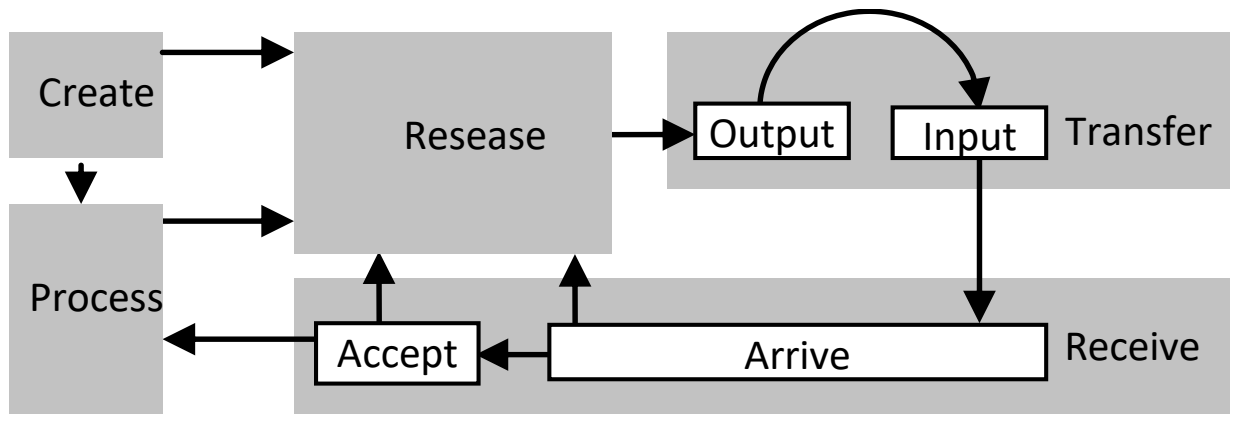

Fig. 2. Thinging machine.

- *Creation*: A new thing is born (created) within a machine. Creation can designate bringing into existence (e.g., ∃ in logic) in the system because what exists is what is found.

Additionally, creation does not necessarily mean existence in the sense of being alive. Creation in M can also refers to atemporal (to be discussed later) appearance or location in the system. It indicates '*there is'* in the system, but not at any particular time.

The TM model also includes the notion of *triggering* that connects two subdiagrams where there is no flow between them. The triggering is represented by dashed arrows in the TM diagram.

The TM modeling establishes three levels of representation:
(1) A **s**tatic structural model, denoted by **S**, is constructed upon the flow of things in five generic actions (i.e., create, process, release, transfer, and receive). The static TM model **S** is a type of the philosophy of presentism (only the present is "real") that is adopted with the twist that the static TM model description contains all presents where Δs (thimacs and subthimacs) *are there* (no temporality). It includes all things to be found at all instances that exist, along with their histories (e.g., gained and lost constituents). The TM dynamic model dissolves these contradictory existences according to time.
(2) A **d**ynamic model, denoted by **D**, identifies hierarchies of events based on five generic events.
(3) A **b**ehavioral model, denoted by **B**, depicts a chronology of events.

### 3.3 Example of Thimacs

According to Waguespack [34], the most concrete concept in the relational (database) paradigm is the tuple, which "corresponds 1-1 with a single concept of reality that it represents. A tuple collects the facts that identify it as a single concept and the facts most closely identified with it." A set of attributes defines the structure of a tuple. Data attributes store data in the tuple.

Fig. 3 shows the model of a tuple as a thimac. In the thimac tuple (circle 1), the attribute values (2) flow to the tuple machine (3), where they are received (4) and processed (5). Such a processing (e.g., concatenation of fields values) triggers creation (6), which is manifested (7) as a whole tuple thing (8) that represents, in Waguespack's [34] words, "a single concept of reality." Now, the tuple can be released, transferred, received, and processed as a thing.

Waguespack [34] defines the concept of a relation (table) that combines a tuple(s) structure and collection. Fig. 4 shows this definition of a table in terms of created tuples (1) that are collected (2 - stored) and then processed (3) to create the table (4).

Fig. 4 gives the impression that only one tuple is created and collected to form a table. To add the concept of a collection, the notion of an event is needed. In TM modeling, an event is built from a change. We will discuss changes and events later in this paper. Fig. 5 shows the model of the event *A tuple flows to a table*. Note that, for simplicity's sake, only the machine side of the event is drawn. The event includes the time and region of the event, in addition to other subthimacs not shown in Fig. 5. To simplify, an event is only represented by its region. Based on such an assumption, Fig. 6 shows the example of a table producing the events $E_1$, $E_2$, $E_3$, and $E_4$.

A chronology of events is generated, as shown in Fig. 7. In Fig. 7, $E_2 \rightarrow E_1$ (Backward arrow) gathers all created tuples as a collection. If all tuples are gathered, then the group of tuples is processed to form a whole called a 'table'. It is like a cowboy collecting wild horses, one by one, to form a *herd* (may be marked) that is taken to the market.

This modeling of tuples and tables demonstrates the meaning of thimacs as a fundamental construct in expressing representations of notions in a TM model. Such a structuring mechanism, which involves **S**, **D**, and **B**, has been used as a base for modeling in many applications, such as network architecture [23], business processes (e.g., monthly salary system) [22], robotic architectural structure [27], security service desks [25], IP phone communication systems [21], and intelligent monitoring systems [29].

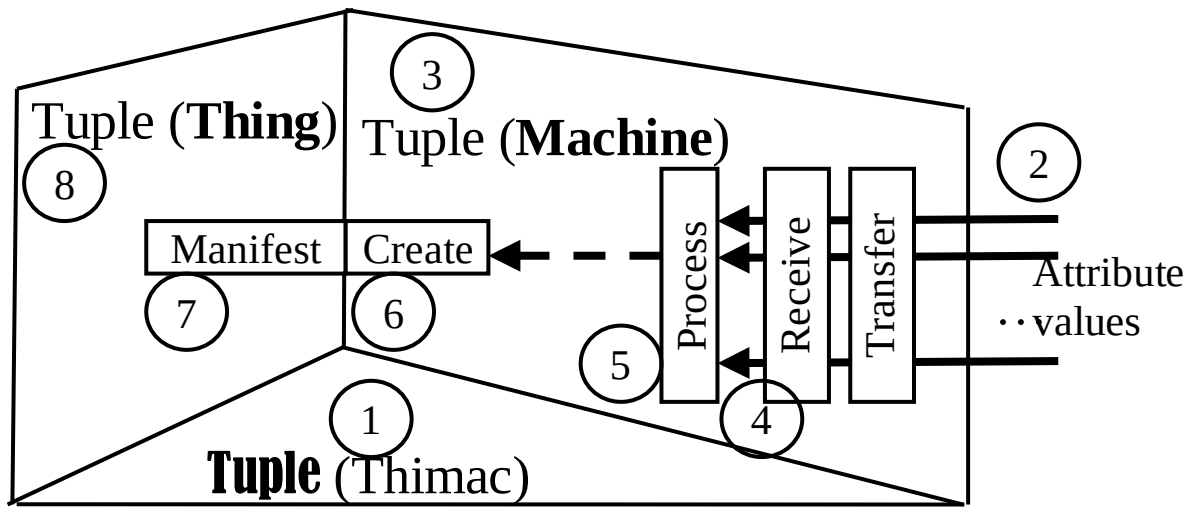

Fig. 3. The thimac tuple has a dual mode of being a thing and a machine.

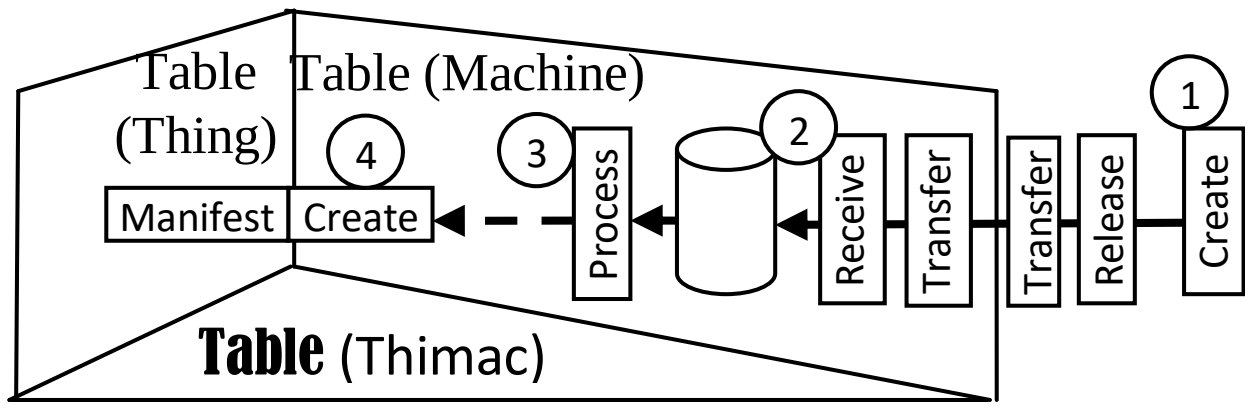

Fig. 4. The thimac tuple is constructed from tuples.

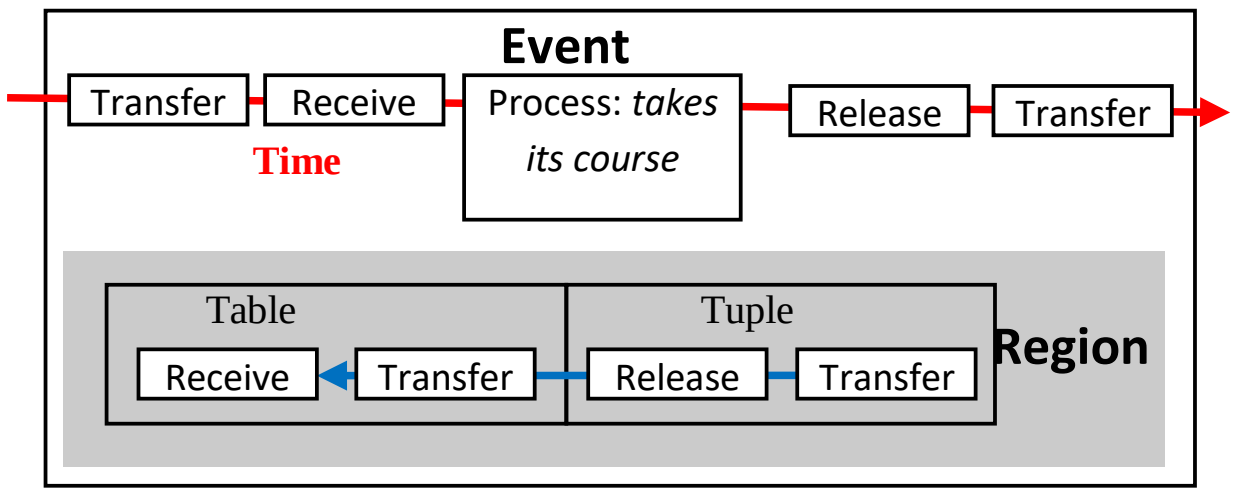

Fig. 5. The event *A tuple flows to a table*.

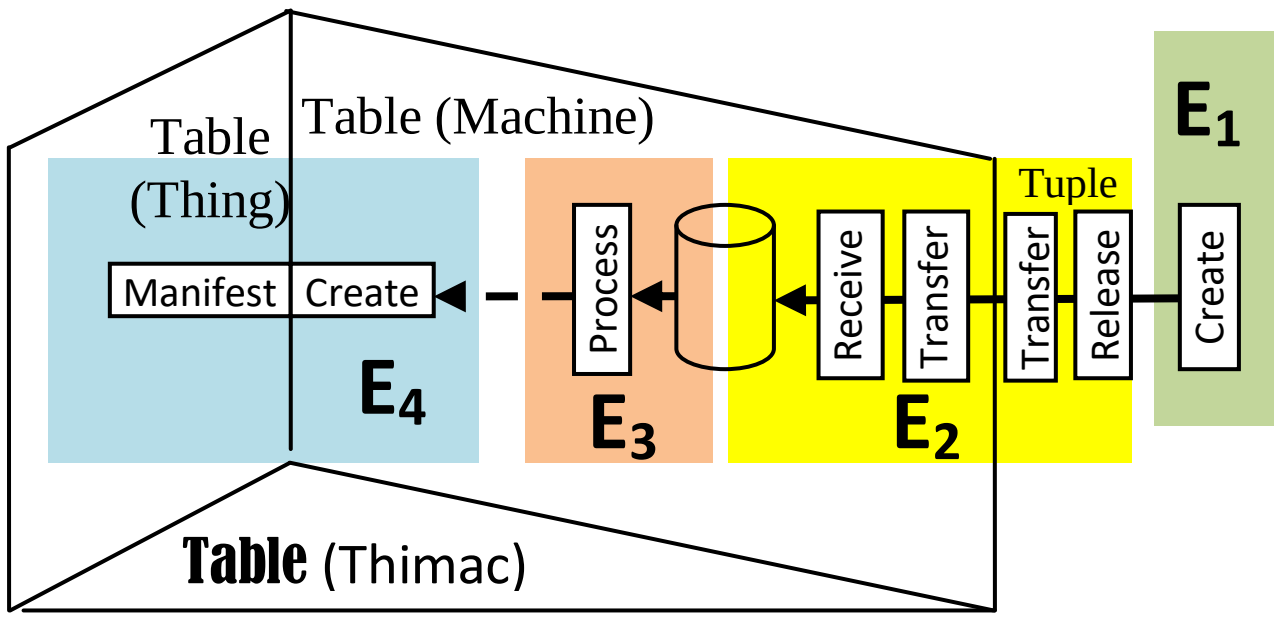

Fig. 6. The thimac table is constructed from tuples.

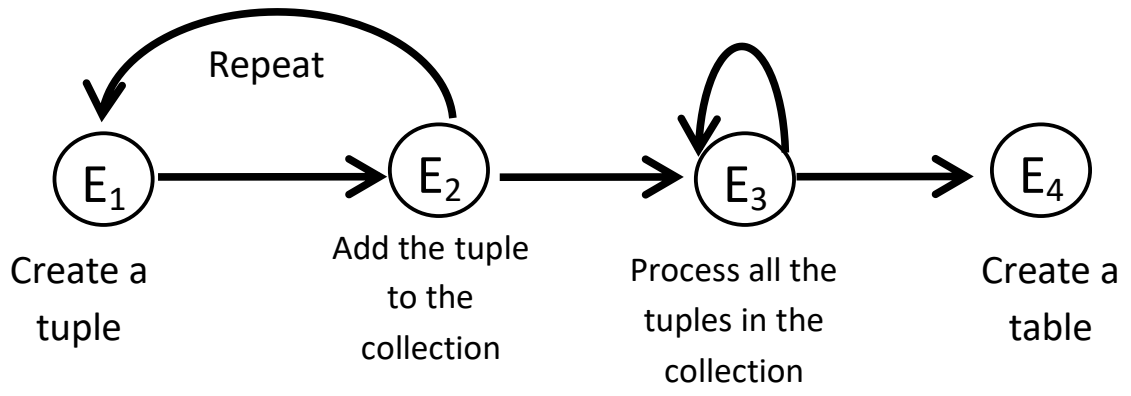

Fig. 7. The model of constructing a table.

## 4. Ontological Problems

In conceptual modeling, ontologies are represented using different descriptions, resulting in a number of semantic interoperability problems among the various ontologies. According to Guizzardi et al. [16], "controlling and defending a particular ontological commitment is essential for the progress of a scientific discipline." Consider the following example problems.

### 4.1 The Heart Transplant and Identity Problem

Rìos [35] discussed the insufficiency of Semantic Web languages to prevent interoperability problems when different ontologies are integrated in a scenario [18]. Rìos [35] presented a fragment of a medical ontology that defines some medically related concepts, such as human organ or human being and surgery room. Rìos [35] describes the following problems: "An application using the medical ontology (Fig. 8) that imports concepts from the legal ontology (Fig. 9) can derive the following wrong information: if a human being receives a heart transplant, he/she becomes a different human being. If the identity of an object is defined by the sum of its parts, then changing one of the parts changes the identity of the object."

The basic problem articulated by this context is how to organize a modeled domain where the main concern lies with assigning entities to positions relative to each other [36]. Ontologies are used as a form of organization, with a static hierarchy of classes similar to the classification of text documents. Kant (1724-1804) suggested that classification is a fundamental aspect of human nature [36]. However, the static description of a system may have nonhierarchical geometrical forms (e.g., circles in astronomy, network in a concept map). As is typical in this discipline, semantic methods are used in this classification.

A TM adds one more ingredient to the organization knowledge base that incorporates processes as entities in the ontology. Process has a static nature and is demonstrated next.

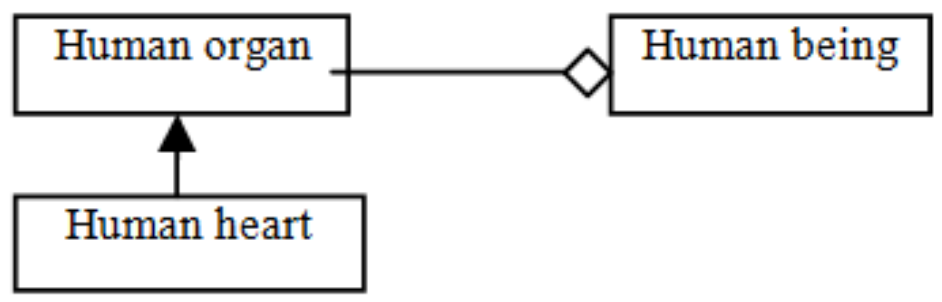

Fig. 8. Fragment of medical ontology (partial, from [35]).

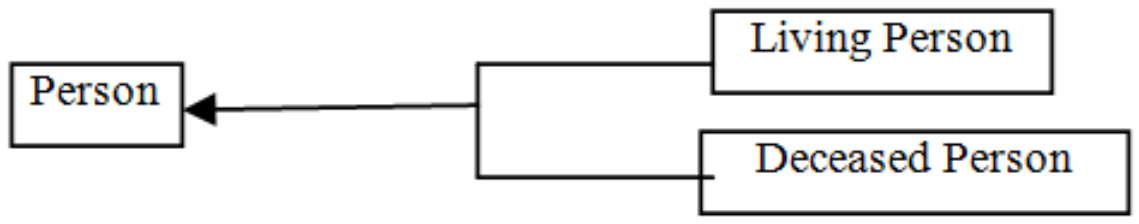

Fig. 9. Fragment of legal ontology (partial, from [35]).

### 4.2 The Static Model 𝕊 of the Heart Transplant Example

Fig. 10 shows the 𝕊 model of this heart transplant. 𝕊 describes the basic "changes" of sending and receiving the heart. In Fig. 10, *create* (circles 1 and 2) indicates that there are (∃) deceased and living human beings with hearts (3 and 4) that appear in the domain of the model.

Such a presence of persons with hearts can be viewed as a "change" in the sense that we start modeling from nothing. As stated previously, *create* indicates that a new thing is born within the machine of the heart transplant thimac. Aristotle stated that substantial *changes* involve a *coming to be*, in contrast to accidental changes, where there is always a substance underlying the change [37].

Next, there is the change of removing the heart from the living person (5 and 6). Last, there is the change of moving the heart of the deceased to become the heart of the living person (7 and 8). The 𝕊 model of Fig. 10 expresses the following: the triggering (9) requires that the moving of the heart from the deceased to the living person can only occur after removing the heart of the living person. The 𝕊 model expresses that *the human being is a thimac that has two subthimacs – living and deceased – each including a human organ called the heart. The heart of the deceased has been moved from the deceased to the living person.*

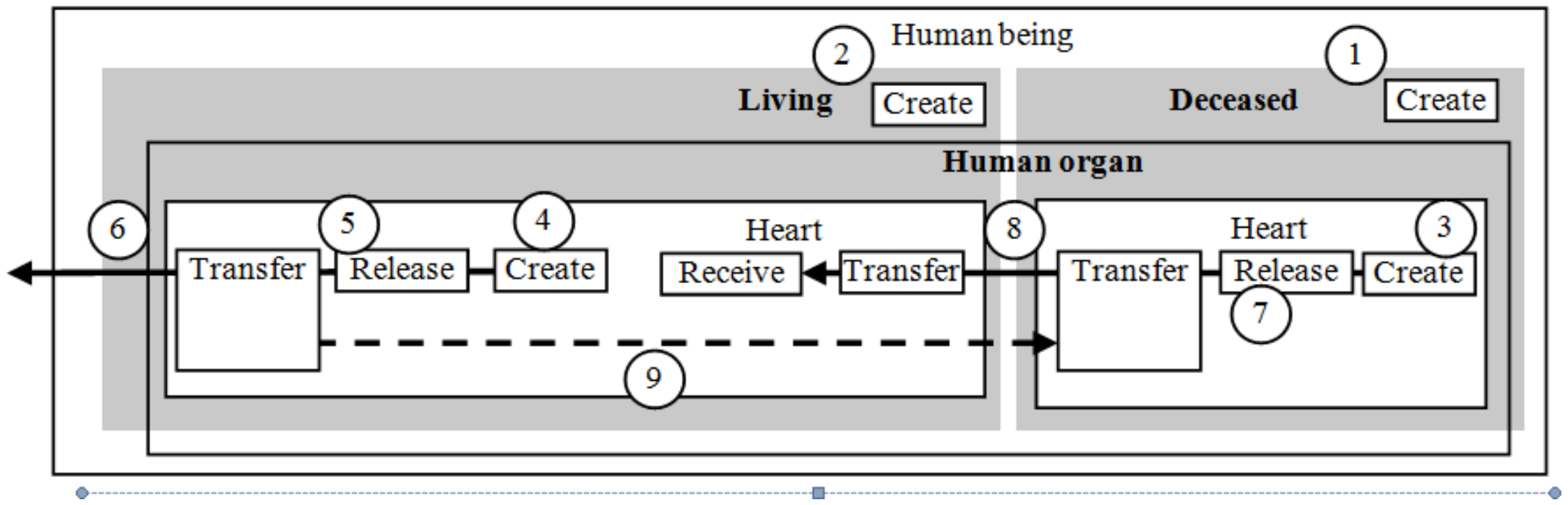

Fig. 10. The 𝕊 TM model of a heart transplant.

The $\mathbb{S}$ model in Fig. 10 includes such nontemporal *changes* as, in the example, the heart flows from deceased to living humans. In general, $\mathbb{S}$ is a world where there are actions without temporality. Thus, creation, processing, release, transference, and receiving are conceived outside of time. These M stages, or a group of them, form the "content" of change. It should be emphasized that $\mathbb{S}$ does not *exist* as a *real* system, but it does embed, simultaneously, potentialities that include all parts of systems' inventory (flows and triggering) stacked on top of each other. In the heart transplant example, the deceased, the living, the removal of the living person's heart, and the transference of the decedent's heart to the living person exist in $\mathbb{S}$. The flows and the triggering may indicate some order of "before" and "after."

$\mathbb{S}$ is a pre-time world; however, this does not imply the absence of the structure of sequentiality. Note that $\mathbb{S}$ is not directly related to ontological studies of the universe wherein a scientist (e.g., Einstein or Gödel) considered the universe to be a timeless phenomenon [38]. Note also that such timeless descriptions were not a new idea, as people, already knew about timeless entities such as numbers, Euclidean triangles, etc.

Change in $\mathbb{S}$ does not involve the passage of time. Aristotle (in *Physics*) argued that change is distinct from time because change occurs at different rates, whereas time does not [39]. In a TM modeling, a change is a region (subdiagram) in $\mathbb{S}$. Nontemporal change "is the difference or nonidentity in the features of things" [39] that can be translated to differences among regions in $\mathbb{S}$.

The $\mathbb{S}$ model is purposely created as a timeless model, in the sense that it includes the past (a human died), the present (there is a deceased and living human), and the future (the heart moves from the deceased to the living humans). Four changes can be identified (see Fig. 11) as follows:

$C_1$: The "existence" of the living person (the change from "there is not" to "there is" – see the first paragraph of the introduction).

$C_2$: The existence of the deceased person.

$C_3$: The removal of the living person's heart.

$C_4$: The moving of the decedent's heart to the living person.

The two-dimensional diagrammatic representation is central as a timeless picture that embeds all changes in the $\mathbb{S}$ model simultaneously in an *atemporal* fashion, except for an order by flow (in general, also, by triggering). $C_1$, $C_2$, $C_3$, and $C_4$ are instances and, in our example, represent (not necessary elementary) units of the appearance in $\mathbb{S}$. $\mathbb{S}$ is neither a space nor a time, but rather a frame for $C_1$, $C_2$, $C_3$, and $C_4$ being relative to each other. This can be generalized for any $\mathbb{S}$ (model of a portion of the world).

### 4.3 Synchronization of Changes

In $\mathbb{S}$, as the frame of changes, order can be imposed to arrive at the atemporal order ("before" and "after"), shown in Fig. 12 and based on the flow from $C_1$ to $C_4$, the flow from $C_2$ to $C_3$, and the triggering from C3 to $C_4$.. Additionally, suppose that there are two heart transplant operations for different persons. Each operation would have its own changes – {$C_1$, $C_2$, $C_3$, $C_4$} and {$C'_1$, $C'_2$, $C'_3$, $C'_4$}. There is no logical reason not to consider {$C_1$, $C_2$, $C'_3$, $C'_4$} or any other mix of changes as an order of changes. However, in our example, this consideration cannot happen because of flows and triggering in the model. The flow (removal) of the heart from the living person to the outside preserves the order {$C_1$, $C_3$} and the flow of the deceased person's heart to the living person {{$C_1$, $C_2$}, $C_4$}. It remains to be seen whether $C_4$ occurs after $C_3$, which is the purpose of triggering. Hence, we can specify the order of changes as shown in Fig. 12. The ordering is captured by the relations of "appears before" or "appears after" or by simultaneous occurrence. In TM modeling, there are no relations – just thimacs. Fig. 13 shows these thimacs (not relations) as $\Delta_1$, $\Delta_2$, and $\Delta_3$.

Accordingly, $\mathbb{S}$ reflects connected chronologies of changes. If it does not, then it produces all possible chronologies. In the heart transplant example, if there is no triggering, then the set of chronologies is {{$C_1$, $C_2$}→$C_3$, {$C_1$, $C_2$}→$C_4$}.

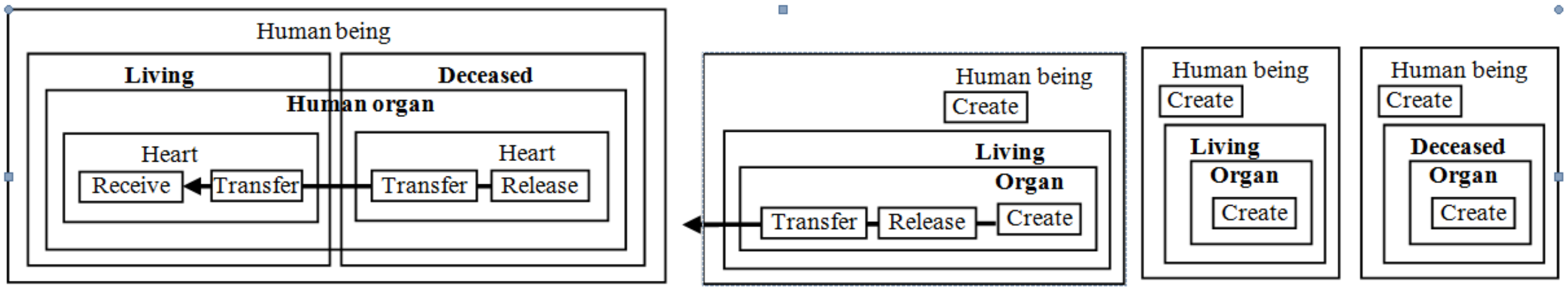

Fig. 11. Dividing the $\mathbb{S}$ model into changes.

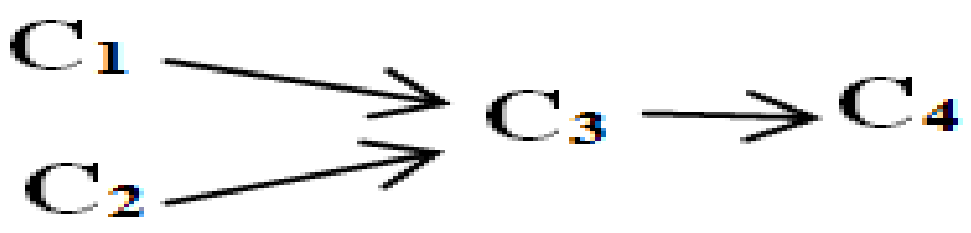

Fig. 12. Two possible orders of instances.

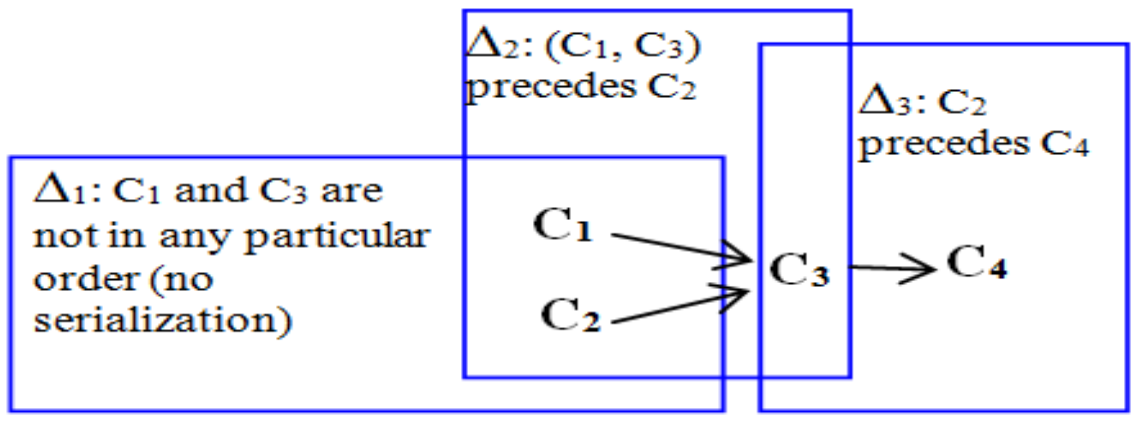

Fig. 13. The changes as thimacs. Note that $C_1$ and $C_3$ are subthimacs.

It is possible that **S** includes several independent components of the chronology of events that are not connected by flows or triggering. We assume that **S** has only one such chronology of events (a graph with only one component).

Fig. 12 (changes and their chronology) reflects the so-called B-series (of time), which is the *series* of all changes ordered in terms of logical relations such as "earlier than," simultaneous, and "later than." The **S** structure covers multiple epochs of change that encroach on each other. Thus, time is not a map, as claimed in B-theory, but a mechanism that "realizes" plots of events. In our example, time realizes the changes in adequate starts and durations (e.g., *moving the deceased person's heart* within an acceptable period after *removing the living person's heart*). The "after" is a change relation, but an "acceptable period" is a time-based imposition.

## 4.4 Transition from Changes to Events

Changes are potential (physical) events. We need a transition from changes in **S** to events in **D**, whereby creation, processing, releasing, transferring and receiving or a group of those stages are placed in time.

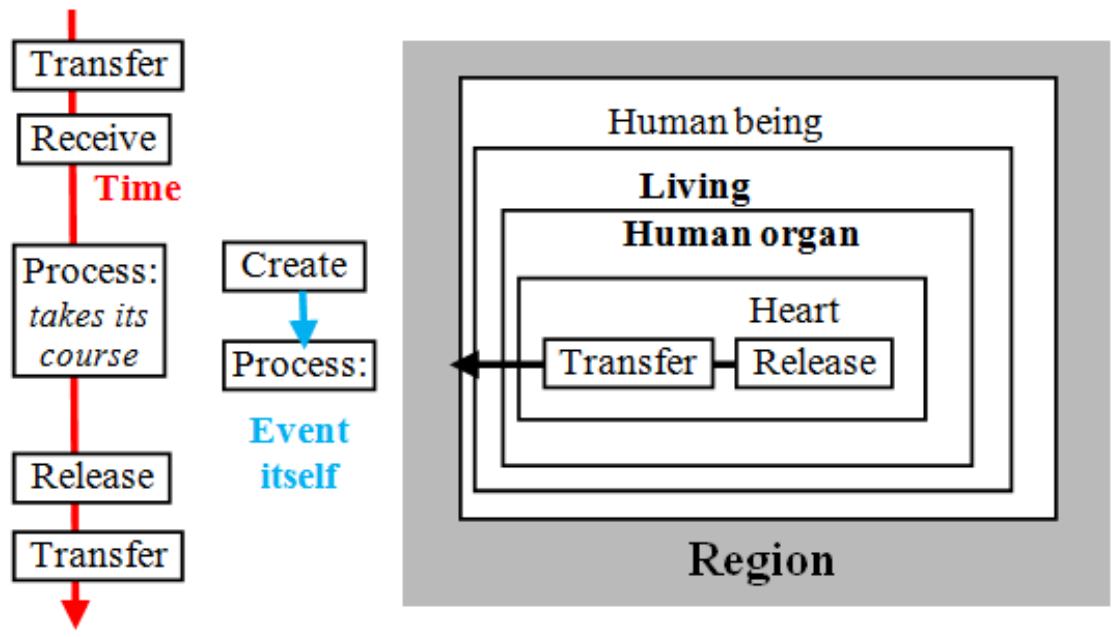

Fig. 14. The event *The heart of the living person is removed*.

In a TM, an event is a Δ. For example, Fig. 14 shows the Δ of the event *The removal of the heart of the living person*.

As stated previously, an event includes the time duration, the region of the change (region of the event), and the event itself, and we usually represent an event by its region. Time brings "practicality" to the model; for example, $C_4$ occurs before $C_2$. (the heart of the living person is removed *before* that of the deceased person is inserted in him) and should happen *within an acceptable period* (wap). Without loss of generality, suppose that $C_1$, $C_2$, $C_3$, and $C_4$ each takes the duration of a century to finish its course, or that differences between their starts are centuries. Still, logically, they satisfy the chronology of events. Hence, the mere insertion of time in a change is a potential for an actual event. By specifying the period and start of each change, such potential situations are eliminated.

Fig. 15 shows the **D** model of the heart transplant case, where each event is assumed to be represented by its region. Hence, the **B** model is produced as a chronology of these events, as shown in the same figure. The realization of the behavior involves an instantiation of physical events. When an instance of **B** (physical events) finishes, the events will form a potential chronology of changes to begin what may be another heart transplant operation. Again, events are thimacs, as shown in Fig. 16.

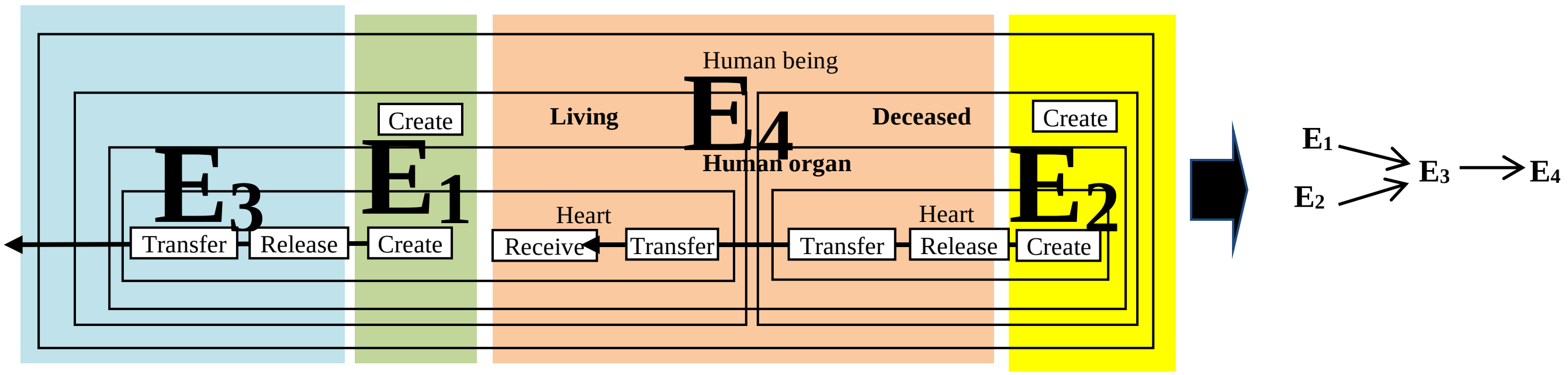

Fig. 15. The **D** model of the heart transplant leads to the **B** model (chronology of events).

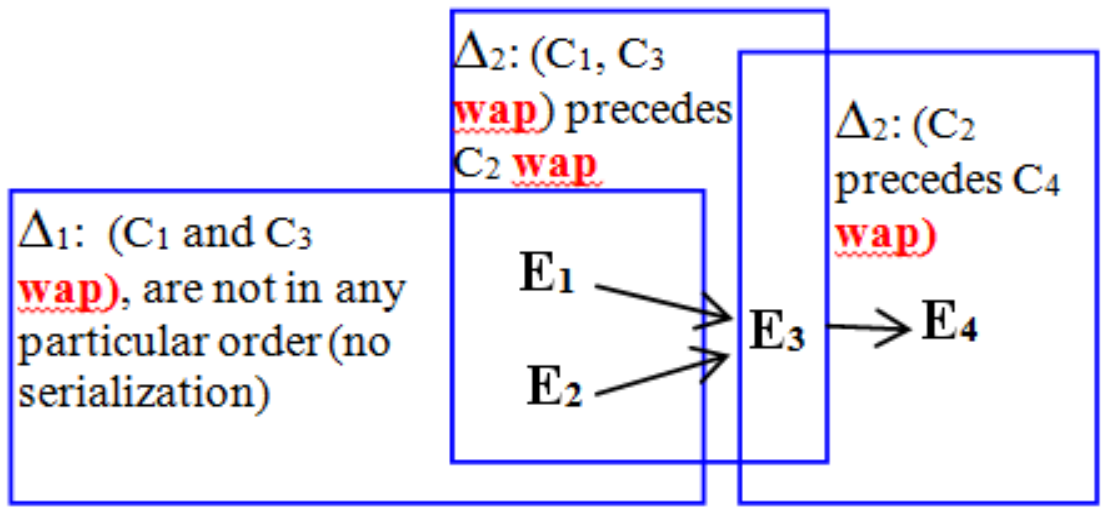

Fig. 16. The events as thimacs make the chronology practical within reasonable time (wap).

Additionally, time is needed in case of repeated changes (e.g., looping). It is clear that a change cannot be repeated consecutively, because the copy of change has the same identical regions and other constitutive parts of the change. Repeated consecutive identical changes represent a single change. The events can be repeated because the repetition has a different time. Repeated events can have identical change. Note that, in 𝕾, a change is defined in terms of a subdiagram. Thus, there is no "empty change." Even if there is waiting, as in some applications, waiting is *created* and may be repeated and thus is a change (subdiagram). This contrasts with the philosophical notion of the passage of time even when nothing happens (e.g., a person feels the passage of time even no change occurs). Time in TM modeling has nothing to do with consciousness.

In TM modeling, because of the isomorphism between the order of changes and the order of events, it is sufficient to use the events diagram. Additionally, it is assumed that the "*within an acceptable period* (wap)" requirement is satisfied.

## 4.5 Flow of Time

We conceptualized the event as a thimac with a time subthimac that includes transfer→ receive→ process→ release→ transfer (see Fig. 14). Thus, apparently, the TM model views time as a thing that flows. The notion of flow of time is not necessary in a TM. According to Williams [40], the flow or passage of time is a sort of illusion. Some researchers think of time as a thing or the "container" or "arena" of all occurrences [41]. Alternatively, a TM may view time as a thimac that processes (unfolding) synchronizations of events, as shown in Fig. 17. The figure has ontological consequences that are not elaborated in this paper.

## 4.6 The Issue of Identity

Rìos [35] stated a problem that, if the identity of an object is defined by the sum of its parts, then changing one of the parts changes the identity of the object, "since if the identity of an object is defined by the sum of its parts, then changing one of the parts changes the identity of the object." However, identity, like everything else, changes in a cumulative way; thus, the identity of an object as the sum of its present parts is a thimac and the "sum" of its parts does change (e.g. create, process, release, transfer and/or receive.). The thimacs retain their identity through change. Rìos's [35] problem originated in the object-oriented ontology assumption. The thimac is a process that "expands" in its totality by incoming and outgoing things; thus, there is no such thing as a complete description of its identity. Locke (1690) asserted that someone can be addressed as the same person if he or she is able to remember previous states in different times and places [42].

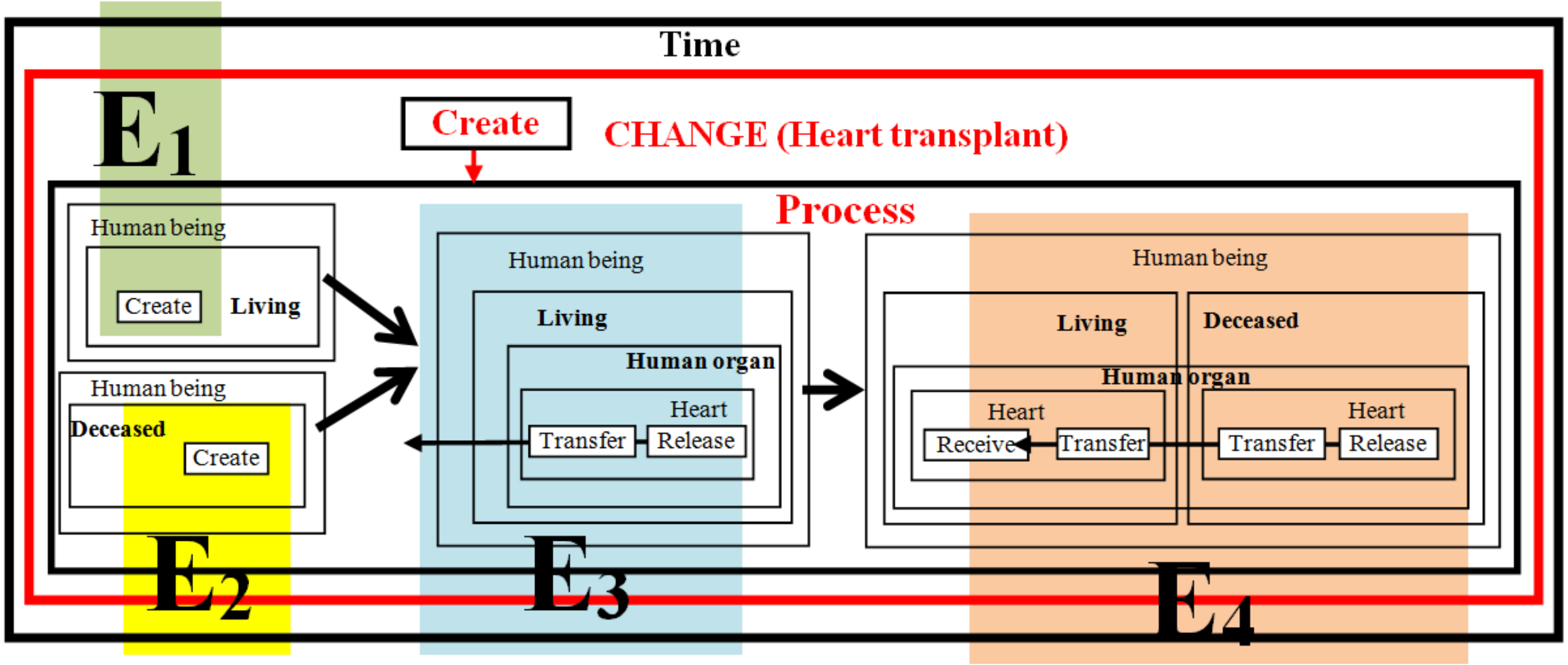

Fig. 17. Time as a thimac that processes changes.

In TM modeling, a person can be addressed as the same person if he or she is the same thimac. Some things and changes or copies of them (but not events) are stored in the thimac and then retrieved and processed. From such a perspective, the replaced part is still part of the identity of the living person in his or her stored portion.

## 5. Ontology Time in SysML

According to Bock and Galey [43], ontology has many applications to engineering, including specifying products in space and time together to enable more reliable modeling. Bock and Galey [43] pointed out that ontology is widely applied in the field of engineering requirements. Specifically, structural requirements can be specified in general systems engineering languages — for example, the Systems Modeling Language (SysML). Structural modeling refers to describing form (e.g., a diagram). The modeling language SysML extends the Unified Modeling Language (UML), "which includes logical interpretations for foundational elements of structural modelling, such as classification, attribution, and composition" [43]. The difficulty in such specification is that requirements in SysML do not completely separate actions from effects. Bock and Galey [43] stated,

> In addition, systems engineering languages do not specify space and time formally enough for automated reasoning and other analysis. They do not usually address space, and extensions to them typically link spatial information to system elements without any other integration, rather than treating spatial extent as an inherent characteristic of system elements enabling them to have spatial relationships.

According to Knorreck et al. [44], the increasing importance of real-time systems has stimulated research work on modeling techniques in such languages as SysML. They asserted, "The use of SysML in verification-centric methods has been hampered by the poor formality of Requirement Diagrams and the lack of powerful property expression language. Thus, UML/SysML profiles commonly require the use of temporal logics" [44].

In this section, the focus is on how to integrate time in modeling, as presented by Bock and Galey [43], in contrast to modeling time in a TM. There is no elaborate discussion of the differences in modeling, but the results of the two approaches, put side by side, do not require much explanation for one to see the distinctive features of each model.

### 5.1 The Car Travel System

Bock and Galey [43] were interested in applying the abstract concepts as classifications of real things according to their characteristics, particularly for space and time. For example, "instead of modelling space as regions and spatial relations on these, a class is introduced for things that exist in space, with relations on them (similarly for time intervals and their relations)" [43]. Fig. 18 shows the class ExistsInSpace for things that take up space, with properties/associations OutsideOf and InsideOf to specify which of the things are outside or inside another, respectively.

Fig. 19 shows the class HappensInTime for things that take time, with the properties/associations HappensBefore and HappensDuring to specify which of the things happen before and at the same time as another, respectively [43]. Fig. 20 shows the TM static model $\mathfrak{S}$ that corresponds to Bock and Galey's [43] model of a car driven by Mary to her garage (Figs. 18 and 19). In the figure, Mary enters her car (circle 1), in which she creates (2) a signal that flows (3) to the car and is there processed (4) to start the car (5). Similarly, she creates a signal (6) that flows to the car (7) and is processed there (8) to trigger the movement of the car. Moving the car triggers (9) the car to move to the traffic area (10), where it is stopped (11). The car is then started (12) to move to the garage (13), where it is stopped (14). There, Mary leaves the car to go to the garage (15).

To inject a time factor and hence events into this description, we must model events. As mentioned previously, an event is a thimac with a time subthimac; for example, Fig. 21 shows the event *Mary enters the car*.

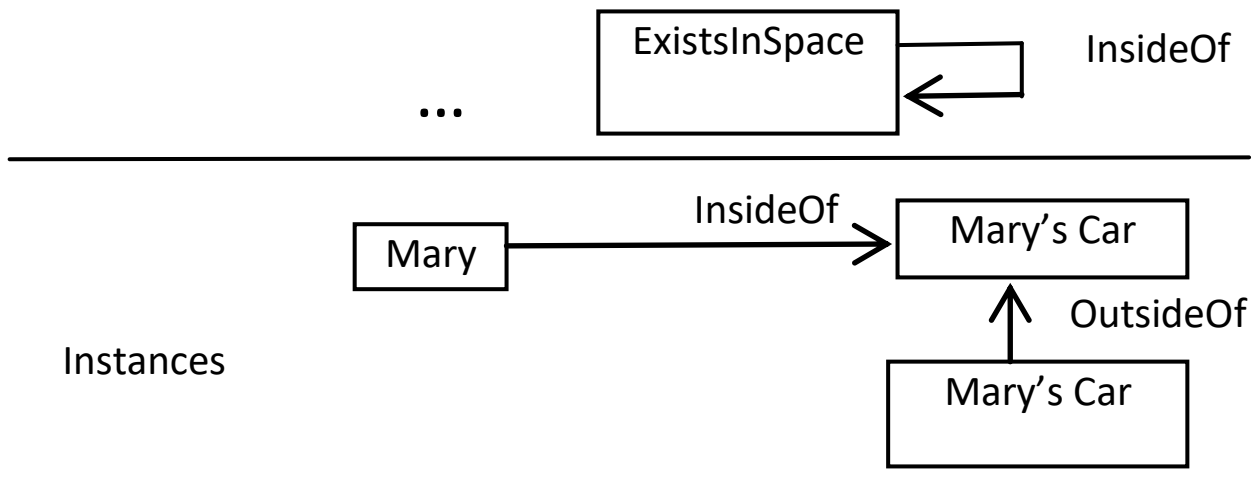

Fig. 18. Example of modeling that involves space and time (redrawn, partial from [43]).

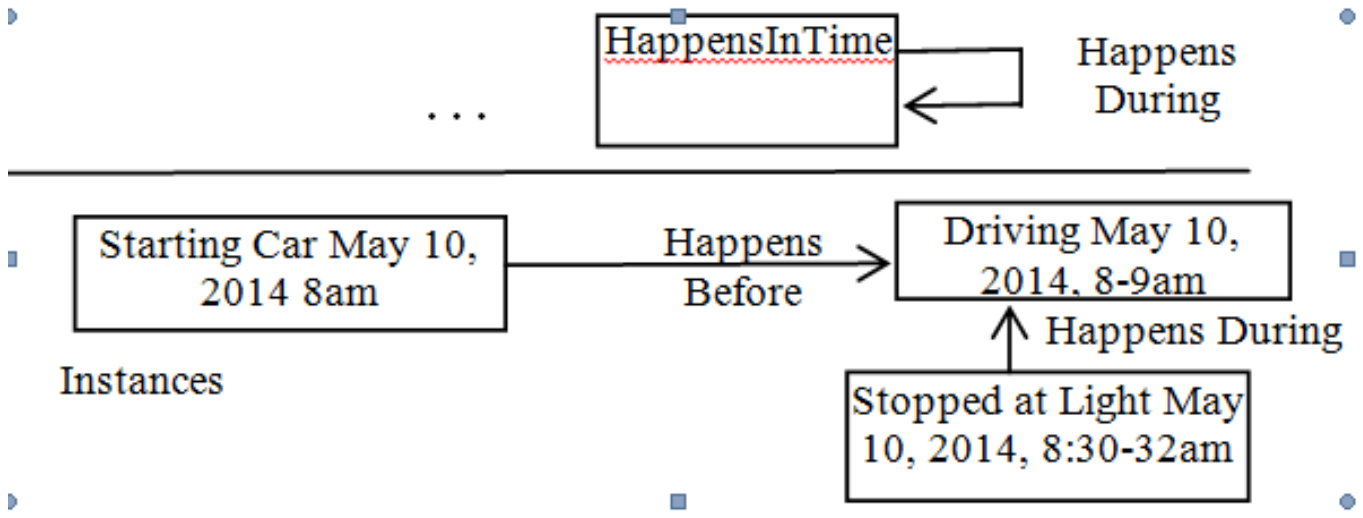

Fig. 19. Example of a class for things that take time (redrawn from [43]).

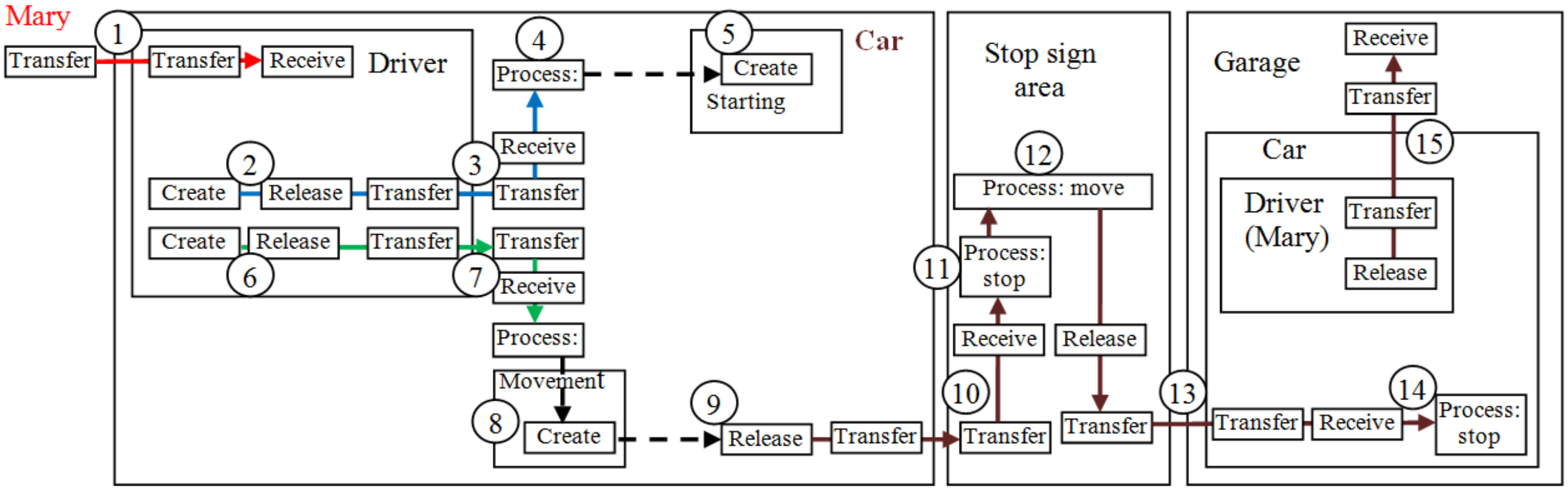

Fig. 20. TM S model of a car driven by Mary to her garage.

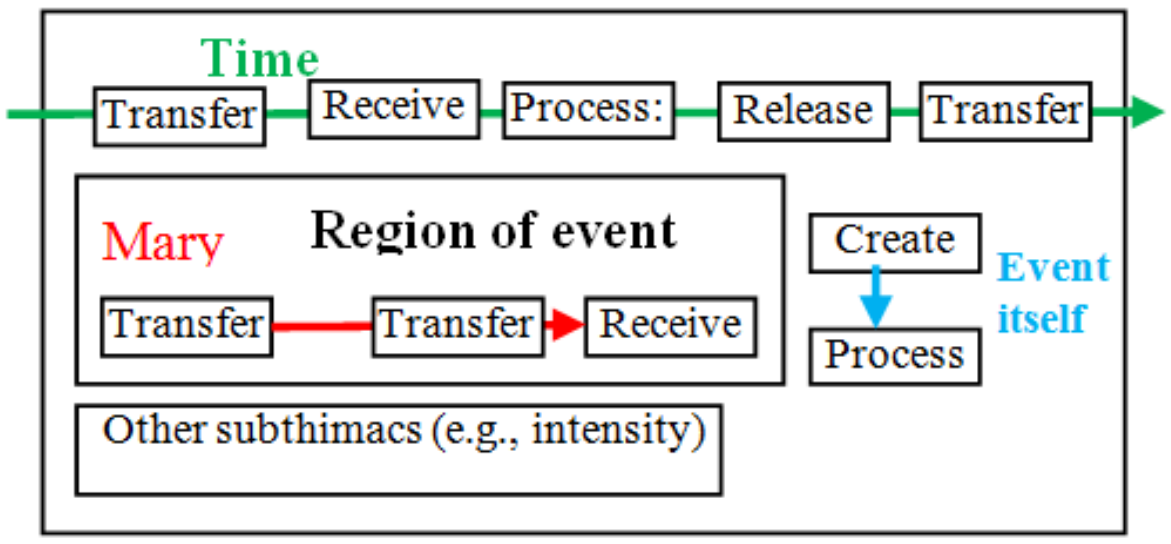

Fig. 21. The event *Mary enters the car*.

The event includes the region (subdiagram of the static model) where the event occurs. For the sake of simplification, we represent events by their regions. Hence, Fig. 22 shows the dynamic model where a set of events are selected. Fig. 23 shows the behavior of the system in terms of the chronology of events.

In the TM modeling, the space and time dimensions are overlaid in the same diagram (S and versions D), Accordingly, events are so-called four-dimensional things that "exist in space-time – as spatio-temporal extents, and in this dimension "things in the past and future exist as well as things in the present" [45]. Thus an event is extended in space as well as time where "the object at a point in time is a temporal part of the whole" [45].

Change is naturally expressed through a four-dimensional classical mereology, which Simons [46], in describesd. A good description of and argument for the 4D paradigm can be found in Sider [47].

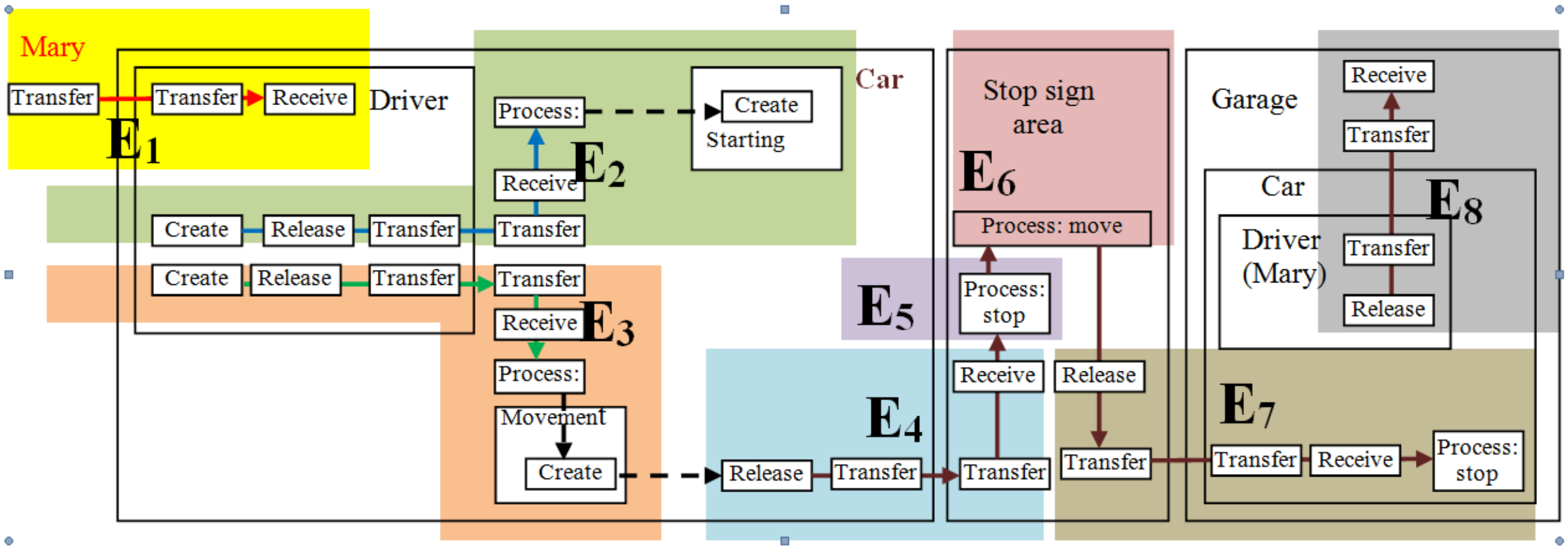

Fig. 22. The S model of a car driven by Mary to her garage.

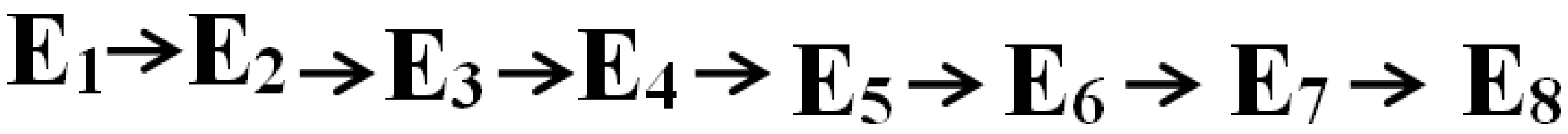

Fig. 23. The B model of a car driven by Mary to her garage.

## 5.2 The Bag Activation System

Ribeiro et al. [48] presented an activity diagram of the air-bag control system to activate the airbag, as shown in Fig. 24, and they listed 15 requirements for the modeling system. We select five of Ribeiro et al.'s [48] requirements related to time, which are as follows:

1. The airbag control system must recognize in a maximum of 5 ms an abrupt deceleration of at least 20 km/h.
2. The airbag control system must recognize the value of the collision impact angle in a maximum of 5 ms.
3. The airbag control system must only activate the airbags if the impact angle is lower than 30 degrees.
4. The airbag control system must only activate the airbags if the collision impact is at frontal movement.
5. The airbag control system must calculate the collision impact angle in at most 5 ms. [48]
6. The airbag control system must recognize frontal movement in at most 5 ms [this requirement is added to align all requirements].

Fig. 25 shows the TM static model **S**. In the figure, the sensors of speed, the angle, and a frontal direction (circles 1 to 3, respectively) send their data to the control system (4, 5, and 6), where the data are processed (7, 8, and 9). If the abovementioned conditions are true (speed of at least 20 km/h, angle lower than 30 degrees, and frontal movement), each triggers activation of the bag (10, 11, and 12). Triggering means sending signals. The thick vertical bar is a simplification notation that all triggering becomes available before activating the bag.

Fig. 26 shows the dynamic model **D**. In **D**, the time that triggers actions appears. There are eight events: $E_1$ through $E_8$.

$E_1$ is the event of the sensor creating the speed data that flow to be processed in the control (circles 1, 2, and 3). Additionally, when these data are created, they trigger registration of the generation time (2 and 3).

Similarly, when the process of the speed data is finished, the finishing time is registered (6 and 7). The generation and finishing times are compared (8), and if the difference is greater than 5 ms (9), a warning is created (10 and 11). This realizes the time constraint on the airbag control system to recognize in a maximum of 5 ms a speed (deceleration) of at least 20 km/h. A similar description can be applied to the angle data (12), $E_2$, and frontal movement data (13), $E_3$.

The event $E_4$ (14) occurs when the following conditions are true: speed of at least 20 km/h, angle lower than 30 degrees, and frontal movement. The event $E_5$ (15) is that the bag is activated. The events $E_6$ (16), $E_7$ (17), and $E_8$ (18) are the warning events. Accordingly, Fig. 27 shows the behavior model **B** of the system. In the figure, $E_1$ occurs repeatedly (reflexive arrow), followed by one of two effects: (i) the bag is activated, or (ii) a warning is created, hence sending the activity back to $E_1$.
Otherwise, nothing happens.
Similar actions are applied to $E_2$ and $E_3$.

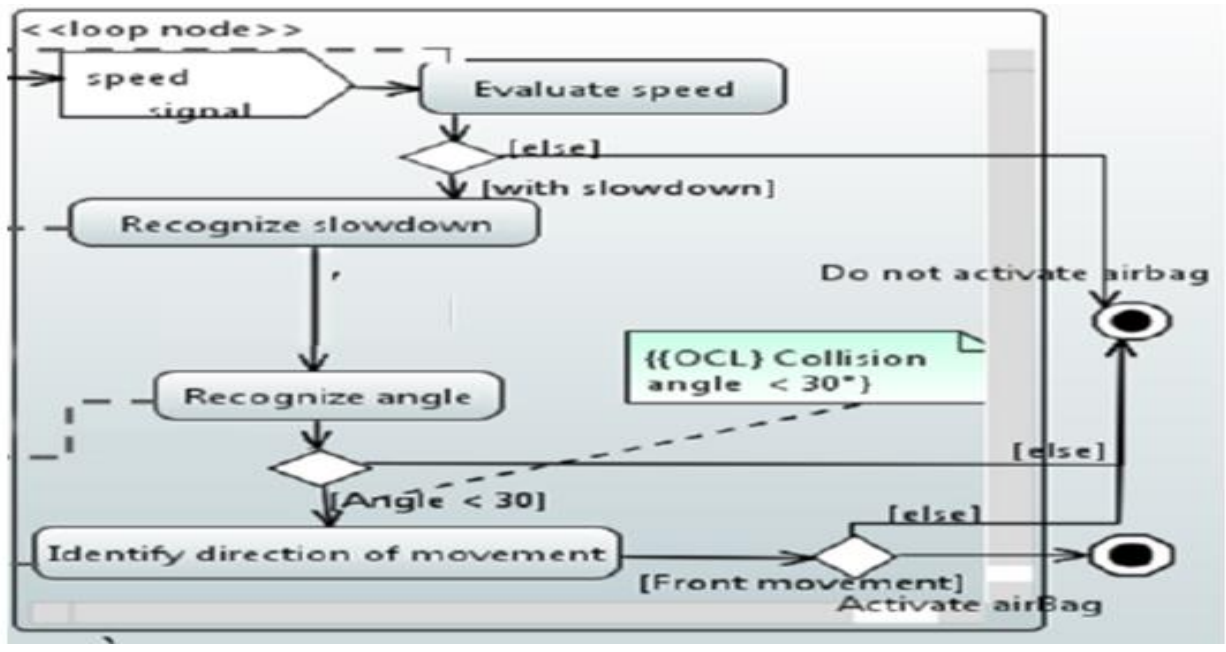

Fig. 24. SysML activity diagram of airbag control system (partial from [48]).

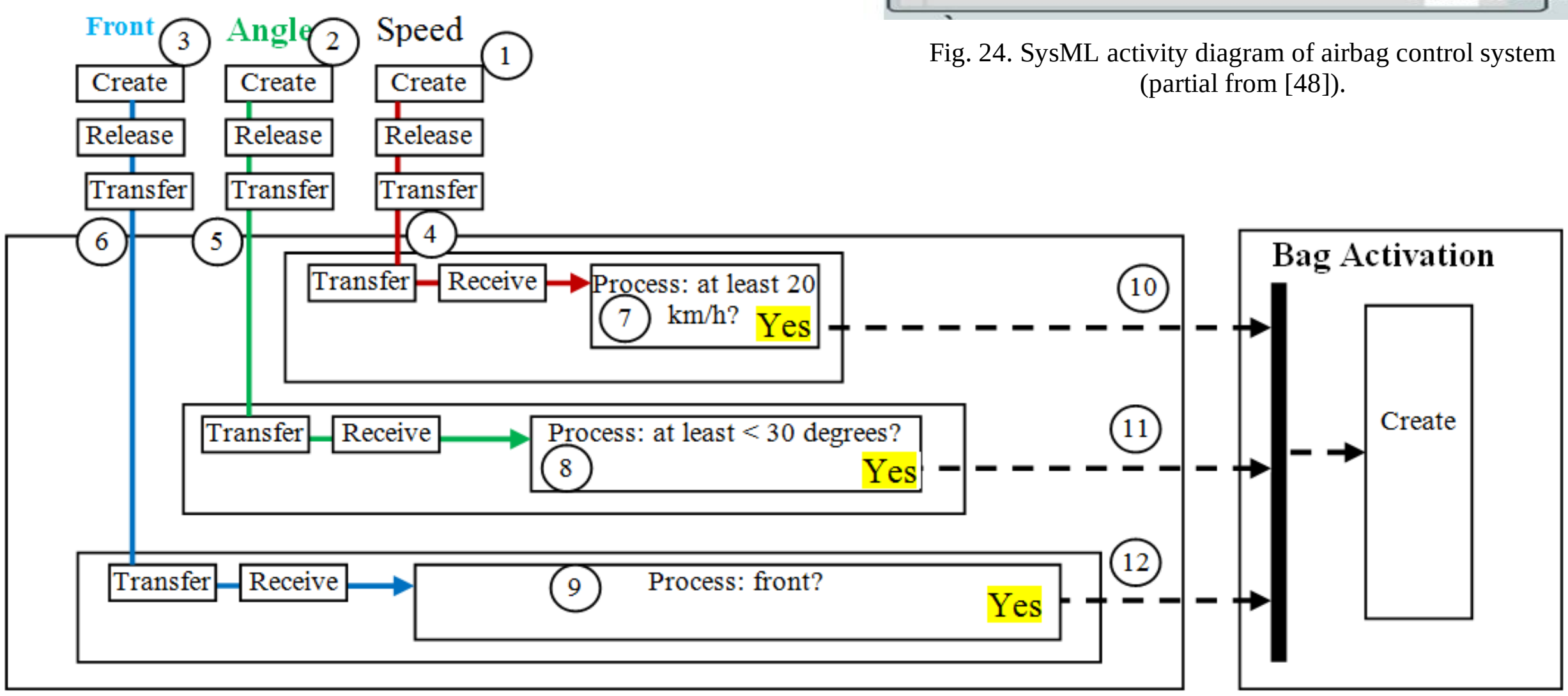

Fig. 25. The **S** model of the bag control system.

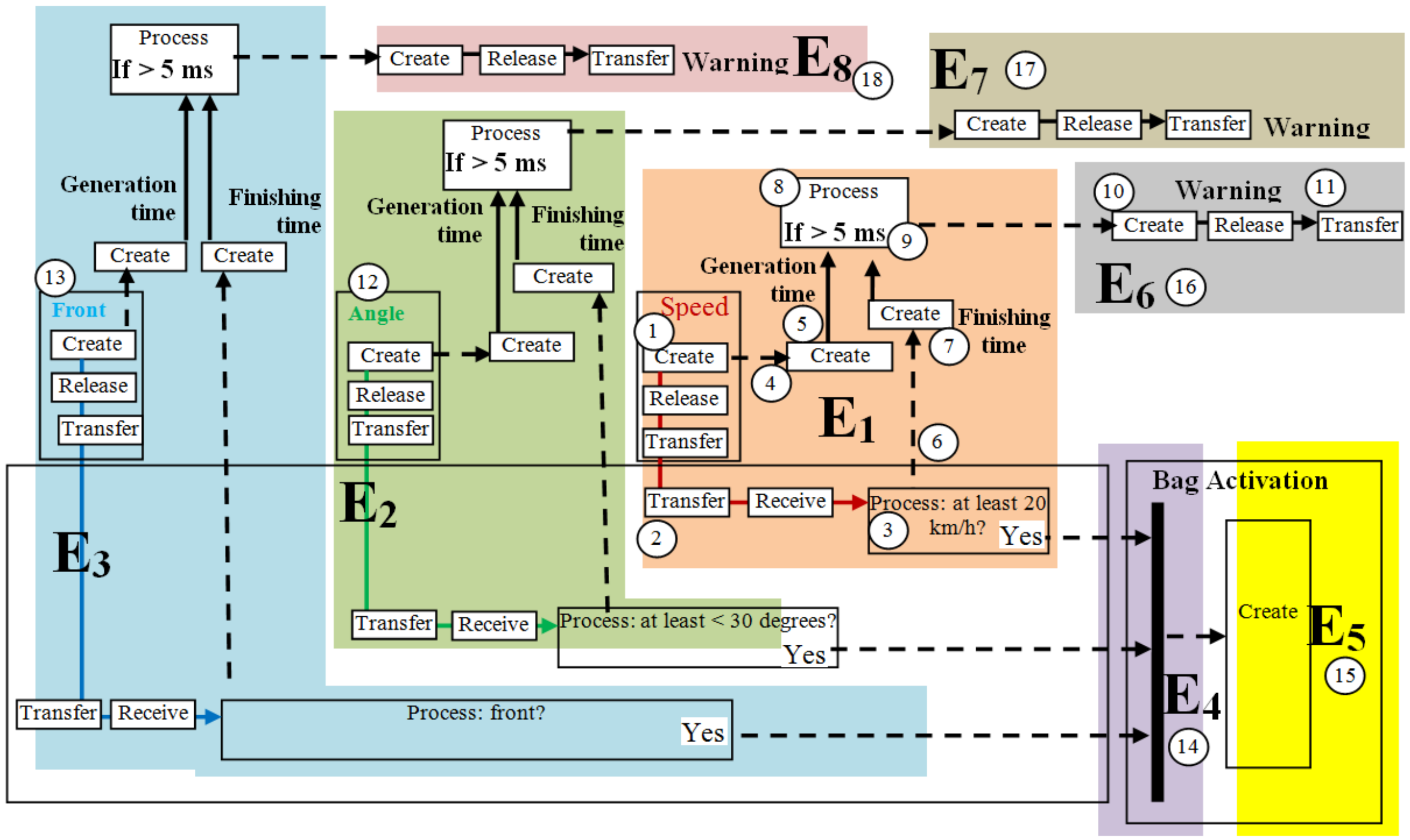

Fig. 26. The D model of the bag system.

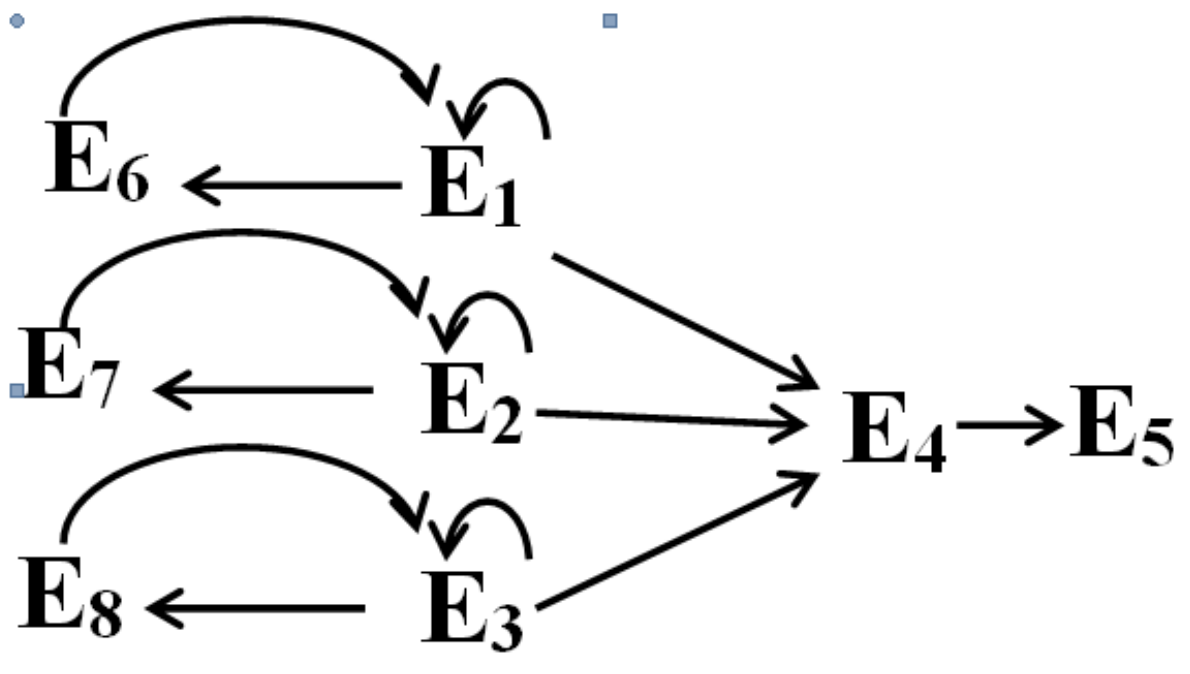

Fig. 27. The B model of the bag control system.

## 6. Conclusion

This paper contributes to establishing a broad ontological foundation for conceptual modeling in the specific domain of reality based on the TM model. Elementary notions of change, events, and time are defined in the context of the TM and applied to the study of selected current ontological difficulties. To demonstrate the viability of the TM approach, examples of modeling in SysML are remodeled.

The results seem to show a clearer and richer representation of the modeled portions of reality. Hence, we claim that TM modeling offers a potential modeling language with the reasonably robust semantics needed in software engineering.

Of course, examining TM features is a continuing process that needs further research. The issue of (visual) diagramming complexity must be addressed. The TM diagram can be simplified by several levels of granularity. Additionally, text-based language for TM modeling can be developed in future research, and more work will be performed on related ontological issues.